\pdfoutput=1
\documentclass[twoside]{article}

\usepackage[accepted]{aistats2022}
\usepackage{natbib} 
\usepackage[utf8]{inputenc} 
\usepackage[T1]{fontenc}    
\usepackage{hyperref}       
\usepackage{url}            
\usepackage{booktabs}       
\usepackage{amsfonts}       
\usepackage{nicefrac}       
\usepackage{microtype}      
\usepackage{amsmath}
\usepackage{subfig}
\usepackage{graphicx}
\usepackage{amssymb}

\global\long\def\jointspace{\mathcal{X}\times\mathcal{X}}
\usepackage[english]{babel}
\usepackage{amsthm}
\usepackage[dvipsnames]{xcolor}
\newtheorem{theorem}{Theorem}

\newtheorem{definition}{Definition}
\theoremstyle{plain}

\begin{document}

\twocolumn[

\aistatstitle{Learning Inconsistent Preferences with Gaussian Processes}

\aistatsauthor{ Siu Lun Chau \And Javier González \And  Dino Sejdinovic }

\aistatsaddress{ University of Oxford \And  Microsoft Research Cambridge \And University of Oxford } ]

\begin{abstract}
We revisit widely used \emph{preferential Gaussian processes} (\textsc{pgp}) by \cite{chu2005preference} and challenge their modelling assumption that imposes rankability of data items via latent utility function values. We propose a generalisation of \textsc{pgp} which can capture more expressive latent preferential structures in the data and thus be used to model inconsistent preferences, i.e. where transitivity is violated, or to discover clusters of comparable items via spectral decomposition of the learned preference functions. We also consider the properties of associated covariance kernel functions and its reproducing kernel Hilbert Space (RKHS), giving a simple construction that satisfies universality in the space of preference functions. Finally, we provide an extensive set of numerical experiments on simulated and real-world datasets showcasing the competitiveness of our proposed method with state-of-the-art. Our experimental findings support the conjecture that violations of rankability are ubiquitous in real-world preferential data. 
\end{abstract}

\section{Introduction}

Data concerning user preferences for items or services is ubiquitous and is often used to detect patterns in user behaviour and to make recommendations. Moreover, these user preferences are often relative (i.e. based on recording choices between a pair of competing items) and may involve an abundance of ranking inconsistencies, e.g. preference of $A$ over $B$, $B$ over $C$, but $C$ over $A$ -- sometimes called a \emph{rock-paper-scissors relation}, and reported, e.g. in mating strategies of certain species \citep{sinervo1996}. Situation like this arises in many domains and is an example of the Condorcet Paradox extensively investigated in social choice theory \citep{gehrlein1983condorcet}. Such inconsistencies may arise due to latent structures determining the criteria for preferences, where different item features may be relevant for making each of these three choices. As an example, consider the case where a cue is present in an item description for $C$, which may be relevant for its comparison to $A$ but not for its comparison to $B$, and that this cue changes the user's criterion when making the choice. Motivated by such inconsistent preferences, we will propose a Gaussian process (GP) model which can capture such latent structures by seamlessly incorporating all the available context information, i.e. sets of item covariates. 

Our main contributions can be summarised as follows:
\begin{enumerate}
    \item We propose a simple generalisation of \textsc{pgp} by \cite{chu2005preference}, allowing to model preferences that do not conform to a consistent ranking. Our method can be integrated directly into many existing probabilistic preference learning algorithms in fields such as rank aggregation \citep{simpson2020scalable}, Bayesian optimisation \citep{gonzalez2017preferential}, duelling bandits \citep{zoghi2015copeland}, recommender systems \citep{nguyen2014gaussian} and reinforcement learning  \citep{zintgraf2018ordered}. 
    
    \item The proposed \textit{Generalised Preferential Gaussian Processes} (\textsc{gpgp}) use \textit{Generalised Preferential Kernels} -- we give a simple construction of these kernels which we prove to satisfy the appropriate notion of universality, i.e. the corresponding RKHS is rich enough to approximate any bounded continuous skew-symmetric function arbitrarily well. While a weaker form of this result has previously appeared in \cite{waegeman2012}, our proof uses different techniques, building on $c_0$-universality notions as developed by \cite{sriperumbudur2011universality}, allowing for more general domains like $\mathbb R^d$.
    
    \item We extend ideas from partial ranking \citep{cheng2012label} and propose a spectral decomposition method to extract \textit{clusters of comparable items} from preferential data using \textsc{gpgp}. This allows us to extract interpretable substructures from a complex network of preferential relationships.
\end{enumerate}

The paper is outlined as follows: in section \ref{sec:bgrnd}, we outline the problem and overview related work. In section 3, we introduce \textsc{gpgp}, describe universality of the corresponding kernel function and how \textsc{gpgp} can be used to uncover clusters of comparable items. Section 4 provides extensive experiments on synthetic and real-world data. Our results improve performance over \textsc{pgp} on all real-world datasets, giving further evidence for ubiquity of inconsistent preferences. We conclude in section 5.

\section{Background}\label{sec:bgrnd}
Assume we would like to choose a data item from domain $\mathcal{X}$. The well established paradigm in this context is \textit{preference learning} (PL), which is concerned with predicting and modeling an order relation on a collection of data items \citep{furnkranz2010preference}. Typical PL models \citep{chu2005preference, d2019ranking, gonzalez2017preferential, houlsby2012collaborative} assume that there is a latent \emph{utility function} $f: \mathcal{X} \rightarrow \mathbb{R}$ to be optimised. We may observe noisy evaluations of $f$ in forms such as item ratings or rankings, but in many cases, an explicit direct feedback from $f$ is scarce or expensive and the quantity of implicit feedback data typically far outweighs the explicit data. Moreover, when the feedback comes from human users, they are better at evaluating relative differences than absolute quantities \citep{kahneman1979interpretation}, and in absence of a reference point explicit feedback may be unreliable and its scale may be ambiguous or difficult to determine. This motivates us to consider the situation where the feedback is \textit{duelling}, i.e. consisting of \textit{binary preferences}. Formally, a pair of items $(x, x') \in \mathcal{X} \times \mathcal{X}$ is presented to the user and we observe a binary outcome which tells us whether $x$ or $x'$ won the duel. For simplicity, we will assume here that no draws are allowed. 

Binary preference data are often represented as Directed Aclycic Graphs (DAGs), where items are denoted as nodes and an edge from node $x \rightarrow x'$ implies that $x$ won the duel over $x'$ \citep{pahikkala2009efficient}. As a result, preference learning can often be seen as learning on DAGs. For example, PageRank \citep{page1999pagerank} can be seen as an Eigenvector centrality measure on a preference graph. For the rest of the paper, we will use the term preference graph and preferential data interchangeably.

One simple model for the duelling feedback is given by 
\begin{equation}
    p(y|(x, x')) = \sigma(yg(x, x')), \quad y \in \{-1, +1\}
    \label{eq: logistic}
\end{equation}
for some $g: \mathcal{X} \times \mathcal{X} \rightarrow \mathbb{R}$, logistic function $\sigma(t) = \frac{1}{1 + e^{-t}}$ and $y = +1$ denoting that $x$ is preferred over $x'$. We note that $g$ must be skew-symmetric, i.e. $g(x, x') = - g(x', x)$ to satisfy the natural condition $p\left(y|(x, x')\right) + p\left(y|(x', x)\right) = 1$ since there are only two outcomes allowed: a win for $x$ or a win for $x'$. Considering general relations on pairs of items in $\mathcal X \times \mathcal X$, \cite{pahikkala2010} term relations which satisfy skew-symmetry \emph{reciprocal}.

An instance of model (\ref{eq: logistic}) is \textit{Preferential Gaussian Process} (\textsc{pgp}) introduced by \cite{chu2005preference}. It is assumed therein that $g$ imposes \emph{rankability} on $\mathcal{X}$. If we define $x \preceq x' \iff g(x, x') \leq 0$, then $\preceq$ is a total order on all of $\mathcal{X}$. This corresponds to writing $g(x, x') = f(x) - f(x')$, where $f$ is the utility function which is determined up to a global shift.  \cite{pahikkala2010} consider a similar notion of a reciprocal relation and term it \emph{weakly ranking representable} when such $f$ exists. In the \textsc{pgp} model, a GP prior is imposed on latent $f$ and the likelihood for a given observation $(x_i, x_j, y_{i,j})$ now becomes
\begin{equation}
    p(y_{i,j}|(x_i, x_j)) = \sigma\left(\left(f(x_i) - f(x_j)\right)y_{i,j}\right).
\end{equation}
Inference on $f$ can then proceed similarly as in GP classification, using methods such as Laplace approximation \citep[Section 3.4]{williams2006gaussian} or variational methods \citep{hensman2015}.

A multitude of probabilistic PL algorithms are developed based on $\textsc{pgp}$. An extension of the model to predict crowd preferences is introduced by \cite{simpson2020scalable}, where a low-rank structure is imposed on the crowd preference matrix and each component is modelled using a GP. On the other hand,  \cite{gonzalez2017preferential} developed preferential Bayesian optimisation to optimise black-box functions where queries only come in the form of duels. \cite{houlsby2012collaborative} incorporated \textsc{pgp} with unsupervised dimensionality reduction for multi-user recommendation systems. Under a similar setting,  \cite{nguyen2014gaussian} applied \textsc{pgp} into a GP factorisation machines to model context-aware recommendations. \textsc{pgp} is also used in the field of reinforcement learning to provide preference elicitation strategies for supporting multi-objective decision making \citep{zintgraf2018ordered}. Finally, one can directly incorporate the learned preference function into learning to rank problems \citep{ailon2010preference}. All models mentioned above assume the data to be perfectly rankable and this is the assumption we challenge in this paper.

Other preference learning models also typically assume data to be rankable and that a well defined utility function exists. Classical examples are \textit{random utility model} \citep{thurstone1994law}, Bradley-Terry-Luce models \citep{bradley1952rank, luce1959possible}, the Thurstone-Mosteller model \citep{mosteller1951experimental} and many of their variants. Non-probabilistic preference models such as SVM-Rank \citep{joachims2009svm}, Serial-Rank \citep{fogel2016spectral}, Sync-Rank \citep{cucuringu2016sync} and SVD-Rank \citep{d2019ranking} also typically assume rankability in their formulations. 

In practice however, total rankability is often too strong of an assumption. There might be many reasons why some ``noisy'' preferences do not conform to a single overall ranking. For example, it is well studied that cognitive biases often lead to inconsistent human preferences in behavioral economics \citep{tversky1992advances}. In fact, not until very recently did the ranking community start to challenge this assumption by proposing quantitative metrics on measuring rankability of duelling data: \cite{anderson2019rankability,cameron2020graph} considered rankability as a metric measuring the difference between the observed preference graph and a perfectly rankable complete dominance graph. This motivates the need to consider a general preference modelling methods without assuming total rankability.

To relax rankability assumptions and thus capture more complex latent structures in preferential data, we will consider a Gaussian process formulation for a general case where no single order can be formed and it is, in particular, possible that transitivity is violated, i.e. $x \preceq x', x' \preceq x''$ but $x'' \preceq x$. We believe that in many cases, such inconsistent relationships are fundamental to the data generating process. In fact, this conjecture is supported by the findings of \cite{zoghi2015copeland} who consider discrete choice (duelling bandits) problem with the application in ranker evaluation for information retrieval. They concluded that the instances where the Condorcet winner (an item which beats all the others with probability larger than $\frac{1}{2}$) does not exist far outweigh those where it does. Since the existence of a single objective function $f$ with a unique global maximum would imply the existence of the Condorcet winner, we see that inconsistent preferences may, in fact, be prevalent in practice. 

A thread of important related work arises in the inference of general (i.e. not necessarily preferential) relations between pairs of data objects \citep{pahikkala2010,waegeman2012} using \emph{frequentist kernel methods}. In particular, \cite{pahikkala2010} similarly emphasise the importance of being able to model \emph{intransitive} reciprocal relationships, motivating it using sports games examples. They also introduce the same kernel function we will consider in this work. \cite{waegeman2012} take this work further, consider more general graded relations, reiterating importance of intransitivity, and study the connections to fuzzy set theory. \cite{waegeman2012} also prove the theoretical result which is a slightly weaker form of our Theorem \ref{thm:ss-univ} on universality. As such, we emphasise that the generalised preferential kernels we will consider are not new, but to the best of our knowledge they have not been used in Gaussian process modelling, nor in discovering richer latent structure behind preferential data, which we propose in this work. There is also work that considers intransitive relations using different types of statistical models -- without using item covariates and operating only on the matrix of match outcomes. For example, \cite{causer2005} extend the classical Bradley-Terry model, while \cite{chen2016} introduce so called Blade-Chest model and discover that substantial intransitivity exists in contexts such as online video gaming data. 

We will in this paper deliberately adopt both Bayesian and frequentist viewpoints to kernel methods. We consider and implement a new Gaussian process framework, generalising \textsc{pgp} of \cite{chu2005preference} which can hence be integrated in many probabilistic preference learning algorithms that build on \textsc{pgp}. But we also study the properties of the RKHSs associated to the corresponding kernel functions, arriving at conclusions essentially equivalent to those in \cite{pahikkala2010, waegeman2012}, although we use different proof techniques which are more grounded in the notions of RKHS universality developed by \cite{sriperumbudur2011universality}, allowing us to consider more general spaces $\mathcal X$ of item covariates. We note that GPs and RKHSs have deep connections, as described in \cite{kanagawa2018gaussian}.

\section{Methodology}


\subsection{Generalised preferential kernels}

Recall that in \textsc{pgp} we express the preference function $g(x, x')$ as $f(x) - f(x
')$ and place a GP prior on $f$. In fact, one can recast the inference solely in terms of $g$ as $f$ directly induces a GP prior on $g$ by linearity. The corresponding covariance kernel $k_E^0$ is then given by
\begin{align}
    k_E^{0}((u, u'), (v, v')) &= \text{cov}(f(u)-f(u'), f(v) - f(v')) \nonumber\\
    &= k(u, v) + k(u', v') \nonumber\\
    &\qquad- k(u, v') - k(u', v),
\end{align}
where the base kernel $k$ is the covariance structure on $f$. \cite{houlsby2012collaborative} called $k_E^{0}$ the \textit{preference kernel}. This reformulation allows us to directly apply many state-of-the-art GP classification methods.  

Now consider a more general case where $g: \mathcal{X} \times \mathcal{X} \rightarrow \mathbb{R}$ corresponds to any skew-symmetric function. We will consider the following skew-symmetric kernel:
\begin{equation}
    k_E((u, u'), (v, v')) = k(u, v)k(u', v') - k(u, v')k(u', v),
\label{eq: sskernel}
\end{equation}
termed \textit{Generalised Preferential Kernel} and the corresponding GP will be called the \textit{Generalised Preferential Gaussian Processes} (\textsc{gpgp}). 

The kernel \eqref{eq: sskernel} is not new and was previously studied by \cite{pahikkala2010, waegeman2012} in their work on intransitive relations, as well as in persistent homology analysis to enforce appropriate symmetry conditions \citep{kwitt2015kerneltopology, reininghaus2015stable}.  In particular, \cite{pahikkala2010} take a feature mapping $\psi$ on $\mathcal X \times \mathcal X$ and ``skew-symmetrise'' it in the following way: $\varphi(x,x')=\psi(x,x')-\psi(x',x)$. Now $\varphi$ and the corresponding kernel can be used to model skew-symmetric functions and, thus, reciprocal relations. In case where $\psi$ corresponds to the Kronecker product kernel $k\otimes k$, this results exactly in \eqref{eq: sskernel}. We give some further details of the feature map view of these kernels in the Appendix. 

One can interpret both $k_E^{0}$ and $k_E$ as kernels between edges in a preference graph. $k_E$ can be extended further to tackle more complex preferential data settings such as \textit{learning from crowd preferences} and \textit{preference learning from distributional data}. We will keep the exposition here simple and a further description of these extensions is included in the Appendix.

For any kernel function $\kappa$, denote its RKHS by $\mathcal H_\kappa$. $\mathcal{H}_{k_E}$ is clearly more expressive than $\mathcal{H}_{k_E^0}$ as it imposes no rankability assumption on its elements. We next consider how expressive $\mathcal{H}_{k_E}$ is, given suitable regularity conditions on $\mathcal{X}$ and $k$. In particular, for \textit{any} skew-symmetric bounded continuous function $g$ on $\jointspace$, can one find a function in $\mathcal{H}_{k_E}$ that arbitrarily well approximates $g$? We define a suitable notion of ss-$c_0$-universality below which allows for a very general domain $\mathcal X$. There are different notions of universality for kernels and we refer the reader to \cite{micchelli2006universal, sriperumbudur2011universality} and references therein for further details.

\begin{definition}[ss-$c_0$-universality] Let $\mathcal X$ be a locally compact Hausdorff space and let $C_{0,ss}(\mathcal{X}\times\mathcal{X})$ be the space of functions $f: \mathcal{X}\times\mathcal{X}\rightarrow \mathbb{R}$ which are continuous, bounded,  skew-symmetric and vanish at infinity. A kernel $k$ is said to be \textit{ss-$c_0$-universal} on $\jointspace$ if and only if $\mathcal{H}_k$ is dense in  $C_{0,ss}(\mathcal{X}\times\mathcal{X})$ w.r.t. the uniform norm.
\end{definition}

We next prove a theorem which allows us to easily construct ss-$c_0$-universal kernels $k_E$ by simply selecting $k$ to be $c_0$-universal \citep{sriperumbudur2011universality}. We note that a weaker form of this result was first proved in \cite[Theorem III.4]{waegeman2012} using different techniques. Our proof (included in Appendix) builds on the notion of $c_0$-universality and its relationship with integrally strictly positive definite kernels developed by \cite{sriperumbudur2011universality}, making the construction applicable to any \emph{locally compact Hausdorff space} $\mathcal X$, whereas \cite{waegeman2012} require \emph{compact metric spaces}, thereby excluding interesting domains such as $\mathbb R^d$ or infinite discrete spaces.

\vspace{0.2em}
\begin{theorem}[ss-$c_0$-universality of $k_E$]
\label{thm:ss-univ}
Assume that the base kernel $k$ is $c_0$-universal on the locally compact Hausdorff space $\mathcal{X}$. Then the generalised preferential kernel $k_E((u, u'), (v, v')) = k(u, v)k(u', v') - k(u, v')k(u', v)$ is ss-$c_0$-universal on $\jointspace$.
\end{theorem}


\subsection{Clusters of comparable items}

Clustering is a popular method to consider latent structures behind preferential data. Many existing methods \citep{cao2012rankcompete, li2018simultaneous, grbovic2013supervised, fogel2016spectral} cluster items based on their similarity devised from the outcomes of matches. For example, in \cite{fogel2016spectral} the authors used a two-hop aggregation method on the preference graph to compute the similarity between two items, i.e. $S_{i,j} = \sum_{k=1}^n y_{i,k}y_{j,k}$. In this work, we consider a different notion of clustering for preferential data, which we term \textit{clusters of comparable items}. In particular, we are interested in discovering groups of items that are comparable and thus rankable within clusters but not across. Cases like this might arise when the pairwise comparison is defined indirectly. For example, product preferences are often deduced using product search histories in e-commerce \citep{karmaker2017application} and products may not always belong to the same categories. 
A related problem is studied in partial rankings \citep{cheng2012label}, where certain pairs of items can be declared as incomparable by thresholding the probabilities of pairwise preferences between items. In contrast to partial rankings though, we do not need to consider individual probabilities, and by clustering the items, all pairings across clusters are declared as incomparable.

Consider a latent preference function $g$ and assume that it belongs to $\mathcal H_{k_E}$. We can associate to $g$ a skew-symmetric Hilbert-Schmidt operator $S_g: \mathcal{H}_k \rightarrow \mathcal{H}_k$ which satisfies

\begin{equation}
   \left\langle k\left(\cdot,x\right), S_g k\left(\cdot,x'\right)\right\rangle_{\mathcal{H}_k} = g\left(x,x'\right).
\end{equation}

For example, if $g(x, x') = f(x) - f(x')$ then $S_g$ is a rank two operator given by $S_g = f \otimes e - e \otimes f$ and $e(x) = 1$ is the constant function. Conversely, if $S_g$ has rank two and one of its top singular functions is constant, a total order can be imposed on $\mathcal{X}$ by the non-constant top singular function. Similar reasoning can also be applied to the match outcomes matrix directly and is the core idea behind SVD-based approaches to ranking \citep{d2019ranking, chau2020spectral}.

In general, however, $S_g$ may have a higher rank. Specifically, in the case of the existence of $L$ clusters of comparable items, $S_g$ can be written as an operator of rank $2L$ given by
\begin{equation}
\label{eq:operator}
    S_g = \sum_{l=1}^L (f_l \otimes e_l - e_l \otimes f_l)
\end{equation}

where $f_l$ is the utility function of the $l$-th cluster and $e_l$ is the $l$-th cluster indicator function, i.e. it equals to 1 if item $x$ belongs to cluster $l$, and $0$ otherwise. 

We are now interested in extracting clusters of comparable items from a fitted function $g$. Assuming \eqref{eq:operator}, the true complete preference matrix $G$ with $G_{i,j} = g(x_i, x_j)$ satisfies
\begin{equation}
    G = \sum_{l=1}^L ({\bf f}_l  {\bf 1}_l^\top - {\bf 1}_l {\bf f}_l^\top).
\end{equation}
${\bf f}_l$ is the vector of evaluations of the $l$-th cluster utility function $f_l$ and ${\bf 1}_l$ is the $l$-th cluster indicator vector, i.e. its $j$-th entry equals to 1 if item $x_j$ belongs to cluster $l$, and $0$ otherwise. 

To recover the clusters, we first estimate the preference matrix $\hat{G}$ using $\textsc{gpgp}$ and treat it as a noisy version of the true low rank matrix $G$. The clusters can then be recovered by applying standard clustering algorithms (e.g. $K$-means) to the data representation given by the top $2L$ singular vectors from $\hat{G}$, analogously to classical spectral clustering. 

\subsection{Data augmentation baseline}
\label{sec:augmentation}
It is simple to extend any classification algorithm to model skew-symmetric duelling preferences using data augmentation, without assuming rankability. One example is to take an observation $(x_i, x_j,y_{ij})$ of the match between $x_i$ and $x_j$, and concatenate the two sets of item covariates in two different orders, as $x_{i,j} = [x_i, x_j]$ and $x_{j,i} = [x_j, x_i]$ and pass them to a classification model with both $x_{i,j}$, $x_{j, i}$ as inputs and $y_{i,j}$ and $y_{j,i}=-y_{i,j}$ as their respective targets. While such data augmentation does encourage skew-symmetry, the resulting function is not guaranteed to be skew-symmetric on all inputs. Skew-symmetry can then be enforced by averaging the model outputs: \begin{eqnarray}
    \hat p(y_{ij}=1|(x_i,x_j))&=&\frac{1}{2}\hat {p}_{\text{cat}}(y_{ij} = 1|x_{i,j}) \nonumber\\&&\;+ \frac{1}{2}\hat {p}_{\text{cat}}(y_{ji} = -1|x_{j,i}),
\end{eqnarray}
where $\hat {p}_{\text {cat}}$ are the probabilities fitted on the concatenated item covariates.
Although this ad-hoc augmentation allows us to relax the rankability assumption in preference learning and is applicable to any models, including GPs, its theoretical justification is questionable, and the additional computational cost due to doubling the data size may be problematic.

We note that another approach applicable to linear models would be to impose skew-symmetry via model coefficients directly, but it is not clear how one might extend it to nonparametric methods such as GPs. We provide further discussion of this line of reasoning with its connection to the feature maps of \textsc{pgp} and \textsc{gpgp} in the Appendix.  

\subsection{Scalability} Since \textsc{gpgp} is formulated on the joint item space $\jointspace$ of pairs of items, computational considerations need to be taken into account. In the worst case scenario, we may be storing and inverting a $\binom{n}{2}$ $\times $ $\binom{n}{2}$ kernel matrix for $n$ items, if a match is played between every pair of items.  This seldom happens in practice, however. In fact, most real-world comparison data is highly sparse, especially if the number of items $n$ is large. Nonetheless, there are a large number of well established ways to scale up GPs that can be readily applied to \textsc{gpgp}, e.g. variational inducing points \citep{hensman2015} or conjugate gradient methods \citep{filippone2015enabling}. In addition, \cite{gardner2018gpytorch} proposed techniques to reduce the asymptotic complexity of exact GP inference from cubic to quadratic. One can also use methods such as KISS-GP \citep{wilson2015kernel} exploiting Kronecker and Toeplitz algebra for further speedups. Kronecker structure of kernel matrices, as well as conjugate gradient methods were also exploited by \cite{pahikkala2013efficient} in the context of regularized least squares with generalised preferential kernel.



\section{Experiments}
Our experiments demonstrate the key aspect of \textsc{gpgp}: the ability to model cyclic and inconsistent preferences from duelling data. In section \ref{sec: cycles}, we study the robustness of \textsc{gpgp} using simulated preferences with different levels of sparsity and inconsistencies. Section \ref{sec: clustering} studies the problem of clusters of comparable items using simulation to further showcase how \textsc{gpgp} can learn complex preferential structures. Finally, we conclude the experiments by testing \textsc{gpgp} against alternative preference prediction methods using 4 real-world datasets with a total of 22 examples. As baselines, we compare \textsc{gpgp} with \textit{Preferential GP} (\textsc{pgp}), \textit{GP with data augmentation}  (\textsc{pair-gp}) and \textit{Logistic Regression with data augmentation}  (\textsc{pair-logreg}). The latter two baselines use a scheme described in \ref{sec:augmentation}.  For all methods involving kernels, we use the Gaussian radial basis function kernel (RBF) $k(x, x') = \exp\big(-\frac{||x - x'||^2}{2\gamma^2}\big)$ and obtain lengthscale $\gamma$ by optimising the evidence lower bound. We use Laplace approximation and conjugate gradient methods for inference in \textsc{gpgp}, \textsc{pgp} and \textsc{pair-gp}.

\subsection{Simulation: Cyclic and inconsistent preferences}
\label{sec: cycles}

\begin{figure*}[!ht]
\centering
 \subfloat[Cyclic and Inconsistent Preference illustration]{%
   \includegraphics[width=0.45\textwidth]{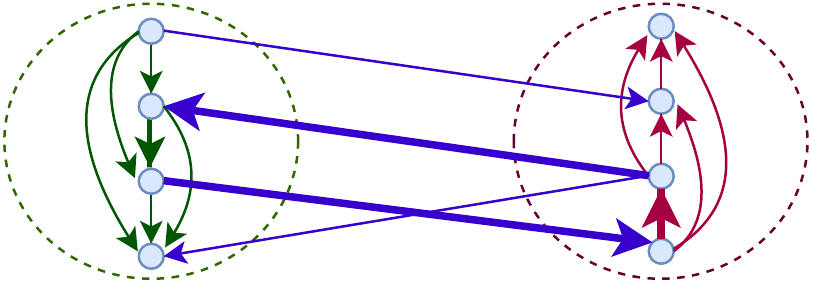}
    \label{fig: cycles}   
 }
 \quad
 \subfloat[\centering{Clusters of comparable items illustration}]{%
   \includegraphics[width=0.45\textwidth]{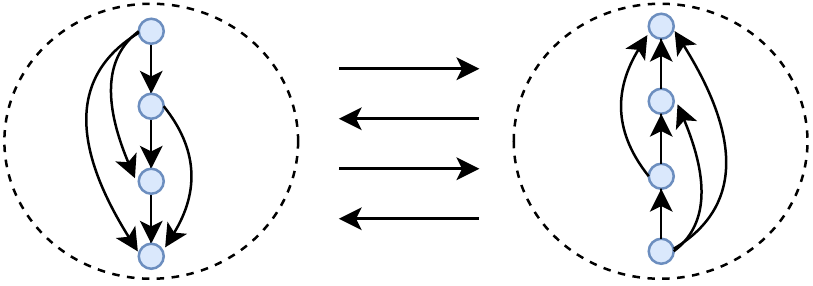}
   \label{fig: clustering}
 }

   \vspace{-2mm}
\caption{(a) Items belongs to different groups and preference between items corresponds to the utility function determined by their latent states (different colors indicate that different utility function is used). Overall preferences exhibit cycles (indicated in bold) (b) Items belongs to different groups and items are rankable within the groups but preferences across groups are random.} 
\label{fig:dummy}
\end{figure*}

\paragraph{Data generation} 
Consider a comparison network with $n$ items and a covariate matrix $X \in \mathbb{R}^{n \times p}$. We assign to each node a latent state $z \in \{1, .., L\}$ and generate a set of utility functions $\{r_{z, z'}\}_{z,z' =1}^L$, i.e. there is a different utility function for each pair $(z,z')$ of latent states. We let $r_{z, z'}(x) = \sum_{j=1}^n\alpha^{z, z'}_j k(x,x_j)$ with 
each vector $\alpha^{z, z'} \stackrel{i.i.d}{\sim} N(0, I_n)$. Comparison between node $i$ and $j$ is then conducted based on the utility selected by their latent states, i.e. $i \preceq j \iff r_{z_i, z_j}(x_i) < r_{z_i, z_j}(x_j)$. This setup brings in cyclic and inconsistent preferences to the overall preference graph. Figure \ref{fig: cycles} provides a visual illustration of the experiment with $L=2$ with a cycle indicated in bold. Different colour of the edges indicates that a different criterion, i.e. utility function, is used in pairwise comparisons.


We simulate a preference graph with $n=30$ players each containing $p=5$ covariates with different level of graph sparsity and number of latent states ($L=1, 2, 5$). Latent states are simulated uniformly. Item features are generated conditionally on latent states with $x|z\sim N(z{\bf 1}, I_5)$, thus allowing the features to encapsulate information about the latent states. We do a $70-30$ train-test-split on the data and repeat the experiments $20$ times.

\begin{figure*}[!ht]
\centering
 \subfloat[L = 1 (Perfectly Rankable)]{%
   \includegraphics[width=0.32\textwidth]{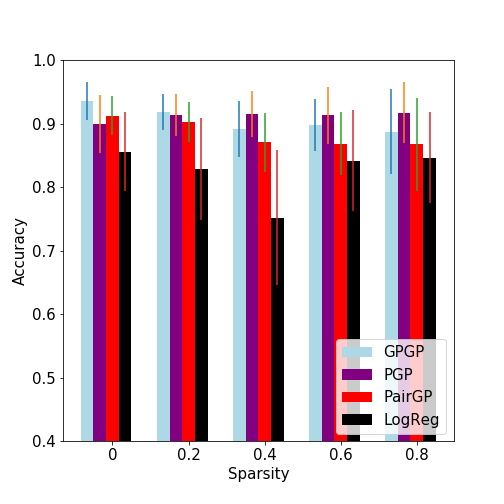}
    \label{fig: NFL_inference}
 } 
 \subfloat[L = 2]{%
   \includegraphics[width=0.32\textwidth]{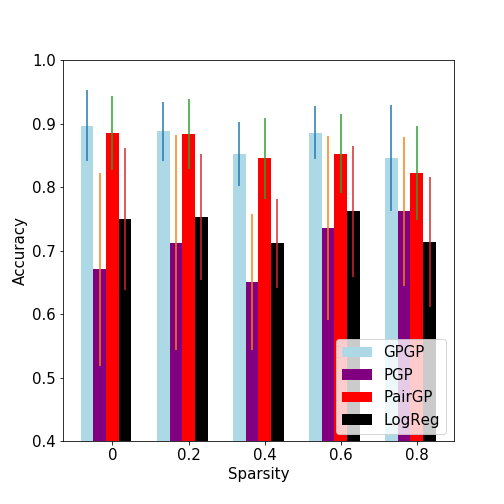}
    \label{fig: NFL_inference_yearly}
 }
 \subfloat[L = 5]{%
   \includegraphics[width=0.32\textwidth]{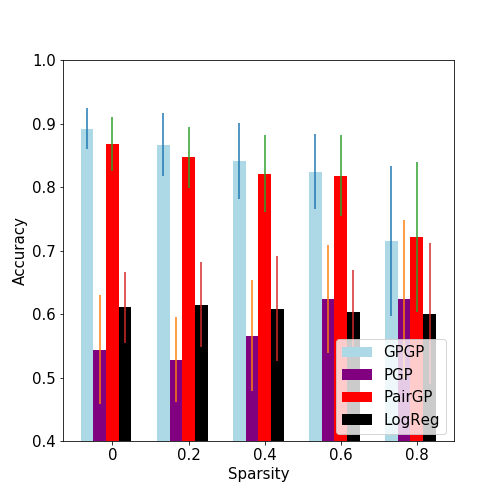}
    \label{fig: NFL_prediction}
 }
 \vspace{-2mm}
\caption{Comparisons of algorithms for simulations at different sparsity and inconsistency level. Accuracies are averaged over 20 runs and error bars of 1 standard deviation are provided.}
\label{fig: simulationcycle}
\end{figure*}

\paragraph{Results}
Figure \ref{fig: simulationcycle} gives the accuracy of \textsc{gpgp} when predicting preferences on held-out data in comparison with baselines. As $L$ increases, we see 
a significant decrease in accuracy for $\textsc{pgp}$ and $\textsc{pair-logreg}$ whereas $\textsc{gpgp}$ and $\textsc{pair-gp}$ performed relatively stable. On average $\textsc{gpgp}$ outperforms the other methods, except in the high sparsity regime with $L=1$, where $\textsc{pgp}$ performed better. In fact, this is not surprising as $L=1$ corresponds to a perfectly rankable duelling problem since there is only one utility function.

\subsection{Simulation: Clusters of comparable items}
\label{sec: clustering}
\paragraph{Data generation} Similar to the setup from section \ref{sec: cycles}, we assign to each data a latent state and match outcomes follow utility functions dependant on these states. However, when comparisons are made across latent groups, the outcome is a Bernoulli($1/2$), independent of all else, due to items being non-comparable. See Figure \ref{fig: clustering} for a visual illustration. We simulate matches between 30 players each containing 5 features with different level of sparsity and number of latent clusters $L = 2, 3$. 

We give three possible approaches of finding the clusters of comparable items,
\begin{enumerate}
    \item \textsc{gpgp-clus}: First recover the latent preference matrix $G$ using \textsc{gpgp}, then run KMeans on the top $2L$ corresponding singular vectors of $G$.
    \item \textsc{pr-clus}: First apply the partial ranking with abstention method from \cite{cheng2012label} to remove non-comparable matches. SVD and KMeans are then applied to the trimmed comparison matrix.
    \item \textsc{svd-clus}: Apply KMeans to the data representation given by the top 2L singular vectors from the comparison graph directly.
\end{enumerate}


We report the proportion of items which are correctly clustered as a metric of performance. We do not include \textsc{pgp-clus} here because \textsc{pgp} performs poorly when there are multiple ranking signals.

\begin{figure*}[!ht]
\centering
 \subfloat[L = 2]{%
   \includegraphics[width=0.32\textwidth]{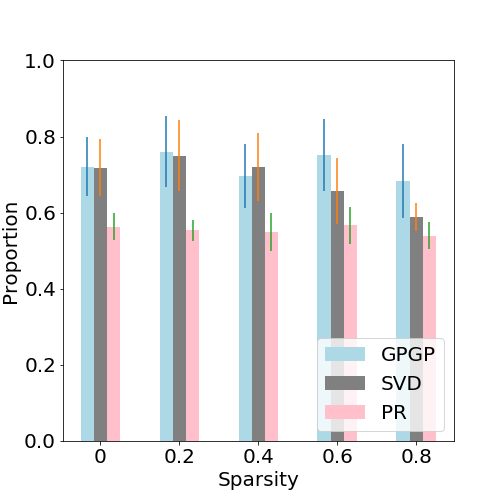}
 } 
 \subfloat[L = 3]{%
   \includegraphics[width=0.32\textwidth]{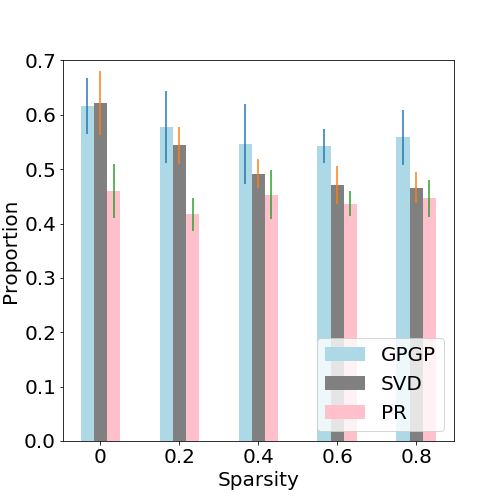}
 }
 \subfloat[]{%
   \includegraphics[width=0.32\textwidth]{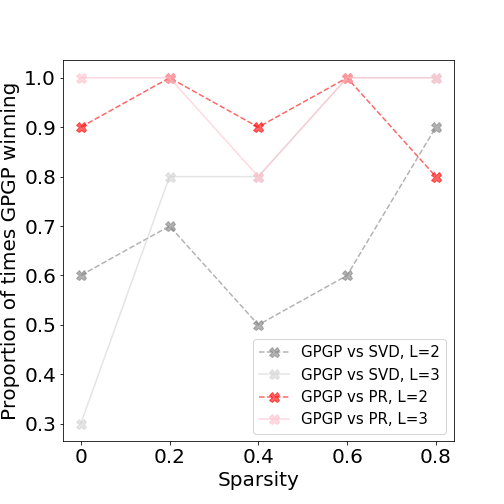}
 }
 \vspace{-2mm}
\caption{(a, b) Comparisons of algorithms for simulations with different number of clusters and sparsity level. Proportion of items correctly clustered are averaged over 20 runs and error bars of 1 standard deviation are provided. (c) Proportion of times \textsc{gpgp} performed better than baselines. }
\label{fig: simulationcluster}
\end{figure*}

\paragraph{Results} Figure \ref{fig: simulationcluster} gives the performance of the methods in recovering clusters of comparable items, comparing the proportion of the items each method clustered correctly. On average \textsc{gpgp-clus} performed better than the rest, except at low sparsity, i.e. dense graphs, where it performed similarly to \textsc{svd-clus}. This is expected as for a highly dense preference graph, modelling with \textsc{gpgp} will not gain further additional information about the overall preference structure. On the other hand, \textsc{pr-clus} performed consistently poorly because it assumes rankability of the data. In other words, it only removes matches that agree with the sole ranking signal the algorithm recovered.

\subsection{Predicting preferences on real data}
\label{sec: real-exp}
We apply \textsc{gpgp} and baselines to a variety of real-world comparison graphs, and measure outcome by their accuracy in predicting preferences on the test set. A 70-30 train-test split is applied to the data over 20 trials. Table \ref{reallifetable} summarises the test results on 4 datasets for preference learning. We report the average network clustering coefficient $C_{avg}$ \citep{saramaki2007generalizations} as a proxy to illustrate how non-rankable the problem is.

\begin{table*}[!ht]
\small
\caption{Test results on the 4 datasets for preference learning. Accuracy averaged over 20 trials is reported along with its standard deviation. $C_{avg}$ is the average clustering coefficient of a comparison graph. The symbol * indicates when the algorithm's accuracy is significantly worse than that of \textsc{gpgp}. Wilcoxon rank-sum test with level 0.05 was used to determine the statistical significance.}
\begin{center}
\scalebox{0.944}{
\begin{tabular}{|l|c|c|c|l|l|l|l|}
\hline
\multicolumn{4}{|l|}{}                                                                    & \multicolumn{4}{c|}{Accuracy (\%)}                                                                                     \\ \hline
DATASET         & {\# Item} & {\# Edge} & $C_{avg}$ & \textsc{gpgp}                         & \textsc{pgp}                            & \textsc{pairgp}                         & \textsc{pairlogreg}                     \\ \hline
Chameleon       & 35                           & 104                          & 0.33       & \textbf{0.78} $\pm$ \textbf{0.06} & 0.51 $\pm$ $0.09^*$ & 0.72 $\pm$ $0.08^*$ & 0.71 $\pm$ $0.08^*$ \\ \hline
Flatlizard      & 77                           & 100                          & 0.07       & \textbf{0.83} $\pm$ \textbf{0.06} & 0.80 $\pm$ 0.09   & 0.78 $\pm$ $0.07^*$ & 0.77 $\pm$ $0.09^*$ \\ \hline
NFL 2000-18 & 32                           & 213x19 yrs                 & 0.54       & 0.59 $\pm$ 0.03 & 0.51 $\pm$ $0.02^*$ & 0.58 $\pm$ 0.03   & \textbf{0.65} $\pm$ \textbf{0.03}   \\ \hline
ArXiv Graph     & 1025                         & 1000                         & 0.11       & \textbf{0.74} $\pm$ \textbf{0.02} & 0.66 $\pm$ $0.03^*$ & 0.70 $\pm$ $0.02^*$ & 0.62 $\pm$ $0.02^*$ \\ \hline
\end{tabular}}
\end{center}
\vspace{1mm}

\label{reallifetable}
\normalsize
\end{table*}

\paragraph{Male Cape Dwarf Chameleons Contest} This data is used in the study by \cite{stuart2006multiple}. Physical measurements are made on 35 male Cape dwarf chameleons, and the results of 104 contests are recorded.  From Table \ref{reallifetable}, we see that \textsc{gpgp} statistically outperformed all baselines. In particular, \textsc{pgp} was the worst performer due to the moderately high clustering coefficient.

\paragraph{Flatlizard Competition} The data is collected at Augrabies Falls National Park (South Africa) in September-October 2002 \citep{WM09}, on the contest performance and background attributes of 77 male flat lizards (\emph{Platysaurus Broadleyi}). The results of 100 contests were recorded, along with 18 physical measurements made on each lizard, such as \textit{weight} and \textit{head size}. This comparison graph has the lowest average clustering coefficient thus is the most rankable compared to the rest. On average \textsc{gpgp} still performed better than \textsc{pgp} but the difference is not statistically significant.

\paragraph{NFL Football 2000-2018} The data contains the outcome of National Football League (NFL) matches during the regular season, for the years 2000 - 2018 \footnote{data collected from nfl.com}. In addition, 256 matches per year between 32 teams, along with 18 performance metrics, such as \textit{yards per game} and \textit{number of fumbles} are recorded. We pick the top 5 informative features by applying the BAHSIC feature selection algorithm \citep{song2012feature} and run the algorithm on each year's comparison graph separately and average the results. In this highly non-rankable $(C_{avg}=0.54)$ problem, \textsc{pair-logreg} outperformed the rest. This is not surprising as the features (e.g. \textit{yards per game}) are expected to be linearly related to the match outcome and a linear model may thus better capture these relationships. Nonetheless,  \textsc{gpgp} still outperformed \textsc{pgp}.

\paragraph{ArXiv Citation Network} The last dataset we use is from the Open Graph Benchmark \citep{hu2020open} arXiv Computer Science papers citation network. Each paper represents a node and an edge from node $i\rightarrow j$ means paper $i$ cited paper $j$. We pick an induced subgraph with 1025 nodes and 1000 edges from the full network. Each node contains a 128-dimensional feature vector obtained by averaging the embedding of words in its title and abstract. Again, we see \textsc{gpgp} performed significantly better than the other algorithms. It is interesting to note that \textsc{pair-logreg} was the worst performing method, indicating that word-embedding features, in contrast to the features from the NFL problem, have a highly non-linear relationship with the match outcome.

\section{Conclusion and Discussion}
\vspace{-0.3cm}
We proposed \textit{Generalised Preferential Gaussian Processes} (\textsc{gpgp}), a new probabilistic model for preferential data. \textsc{gpgp} relaxed the rankability assumption and comes with a strong theoretical justification in terms of universality of the corresponding kernel function. It can be readily integrated into many existing preference learning algorithms that are based on \textsc{pgp}. Experimental results on simulations and real-world datasets show the superior performance in comparison to \textsc{pgp}, the latter demonstrating the prevalence of inconsistent preferences and the need for relaxing the rankability assumptions in practice. We demonstrated how \textsc{gpgp} can be used to solve a specific problem which goes beyond rankability, i.e. recovering clusters of comparable items. A number of other problems which similarly involve more complex preferential structures can be studied based on the proposed framework. 

Relaxing rankability allows to investigate latent structures influencing preferences, including the case where preferences are inconsistent, cyclical or when many items are simply not comparable to each other. Building on the existing preferential Gaussian Process (\textsc{pgp}) model, our approach introduces additional flexibility but preserves the advantages of having a Bayesian probabilistic model and faithful uncertainty quantification. The algorithms we proposed may enable more robust and customised recommendations to users in recommender systems and information retrieval. It is also envisaged that our work will find applications in A/B testing, gaming systems, and Bayesian optimisation with implicit or relative feedback.

Digital trails such as web searches and purchase patterns are often collected for targeted recommendations. It is worth noting that these features might include sensitive personal information and utilising them without careful consideration might be unethical. Therefore, an important practical research direction will be to consider combining \textsc{gpgp} with algorithmic fairness approaches applicable to kernel methods and GPs \citep{li2019kernel}, or to use differentially private mechanisms for GPs \citep{Smith-dpgp16}. 



\normalsize


\end{document}